\documentclass[runningheads,a4paper]{llncs}
\usepackage{amssymb}
\usepackage{amsmath}
\usepackage{amsfonts}
\usepackage{algorithm}
\usepackage{makeidx}
\usepackage{graphicx}
\usepackage{epstopdf}
\usepackage{diagbox}
\usepackage{booktabs}
\usepackage[usenames,dvipsnames]{color}
\usepackage{subcaption}
\usepackage{caption}
\usepackage{multirow}
\usepackage{url}

\usepackage[misc]{ifsym}

\begin{document}

\mainmatter  


\title{Heterogeneous Graph Convolutional Neural Network via Hodge-Laplacian for Brain Functional Data}

\titlerunning{HL-HGCNN}

%
\author{Jinghan Huang$^{1}$ 
\and Moo K. Chung$^{2}$
\and Anqi Qiu$^{1,3-6}$\textsuperscript{(\Letter)}
}

\institute{$^{1}$ Department of Biomedical Engineering, National University of Singapore\\
\email{bieqa@nus.edu.sg} \\
$^{2}$  Department of Biostatistics and Medical Informatics, the University of Wisconsin-Madison, Wisconsin,  USA\\
$^{3}$ NUS (Suzhou) Research Institute,  National University of Singapore,  China \\
$^{4}$ Institute of Data Science, National University of Singapore, Singapore  \\ 
$^{5}$  The N.1 Institute for Health, National University of Singapore, Singapore \\
$^{6}$ Department of Biomedical Engineering, The Johns Hopkins University,  USA 
}

\authorrunning{J. Huang et al.}

\maketitle
\vspace{-0.2cm}
\begin{abstract}
This study proposes a novel heterogeneous graph convolutional neural network (HGCNN) to handle complex brain fMRI data at regional and across-region levels. We introduce a generic formulation of spectral filters on heterogeneous graphs by introducing the $k-th$ Hodge-Laplacian (HL) operator. In particular, we propose Laguerre polynomial approximations of HL spectral filters and prove that their spatial localization on graphs is related to the polynomial order. Furthermore, based on the bijection property of boundary operators on simplex graphs, we introduce a generic topological graph pooling (TGPool) method that can be used at any dimensional simplices.  This study designs HL-node, HL-edge, and HL-HGCNN neural networks to learn signal representation at a graph node, edge levels, and both, respectively. Our experiments employ fMRI from the Adolescent Brain Cognitive Development (ABCD; n=7693) to predict general intelligence. Our results demonstrate the advantage of the HL-edge network over the HL-node network when functional brain connectivity is considered as features. The HL-HGCNN outperforms the state-of-the-art graph neural networks (GNNs) approaches, such as GAT, BrainGNN, dGCN, BrainNetCNN, and Hypergraph NN. The functional connectivity features learned from the HL-HGCNN are meaningful in interpreting neural circuits related to general intelligence.

\end{abstract}

\vspace{-0.8cm}
\section{Introduction}

Functional magnetic resonance imaging (fMRI) is one of the non-invasive imaging techniques to measure blood oxygen level dependency (BOLD) signals \cite{glover2011overview}. The fluctuation of fMRI time series signals can characterize brain activity. The synchronization of fMRI time series describes the functional connectivity among brain regions for understanding brain functional organization. 

There has been a growing interest in using graph neural network (GNN) to learn the features of fMRI time series and functional connectivity that are relevant to cognition or mental disorders \cite{li2020braingnn,zhao2022dynamic}.

GNN often considers a brain functional network as a binary undirected graph, where nodes are brain regions, and edges denote which two brain regions are functionally connected. Functional time series, functional connectivity, or graph metrics (i.e., degree, strength, clustering coefficients, participation, etc.) are defined as a multi-dimensional signal at each node. A substantial body of research implements an convolutional operator over nodes of a graph in the spatial domain, where the convolutional operator computes the fMRI feature of each node via aggregating the features from its neighborhood nodes \cite{zhao2022dynamic,li2020braingnn}. Various forms of GNN with spatial graph convolution are implemented via 1) introducing an attention mechanism to graph convolution by specifying different weights to different nodes in a neighborhood (GAT,  \cite{hu2021gat}); 2) introducing a clustering-based embedding method over all the nodes and pooling the graph based on the importance of nodes (BrainGNN, \cite{li2020braingnn}); 3) designing an edge-weight-aware message passing mechanism  \cite{cui2022interpretable}; 4) training dynamic brain functional networks based on updated nodes’ features (dGCN, \cite{zhao2022dynamic}).   BrainGNN and dGCN achieve superior performance on Autism Spectrum Disorder (ASD) \cite{li2020braingnn} and attention deficit hyperactivity disorder (ADHD) classification \cite{zhao2022dynamic}. Graph convolution has also been solved in the spectral domain via the graph Laplacian \cite{bruna2013spectral}. For the sake of computational efficiency when graphs are large,  the Chebyshev polynomials and other polynomials were introduced to approximate spectral filters for GNN \cite{defferrard2016convolutional,huang2021revisiting}.  
For large graphs, the spectral graph convolution with a polynomial approximation is computationally efficient and spatially localized \cite{huang2021revisiting}.

Despite the success of the GNN techniques on cognitive prediction and disease classification \cite{li2020braingnn,zhao2022dynamic}, the graph convolution aggregates brain functional features only over nodes and updates features for each node of the graph. Nevertheless, signal transfer from one brain region to another is through their connection, which can, to some extent, be characterized by their functional connectivity. The strength of the connectivity determines which edges signals pass through. Therefore, there is a need for heterogeneous graphs with different types of information attached to nodes, such as functional time series and node efficiency, and edges, such as functional connectivity and path length. 

Lately, a few studies have focused on smoothing signals through the topological connection of edges \cite{jo2021edge,jiang2019censnet}. Kawahara et al. \cite{kawahara2017brainnetcnn} proposed BrainNetCNN to aggregate brain functional connectivities among edges. However, brain functional connectivity matrices at each layer are no longer symmetric as the construction nature of the brain functional network.  Jo et al. \cite{jo2021edge} employed a dual graph with the switch of nodes and edges of an original graph so that the GNN approaches described above can be applied (Hypergraph NN).  But, the dual graph normally increases the dimensionality of a graph. To overcome this, Jo et al. \cite{jo2021edge} only considered important edges. Similarly, Jiang et al. \cite{jiang2019censnet} introduced convolution with edge-node switching that embeds both nodes and edges to a latent feature space. When graphs are not sparse, the computation of this approach can be intensive.  The above-mentioned edge-node switching based model achieved great success on social network and molecular science \cite{jo2021edge,jiang2019censnet},  suggesting that GNN approaches on graph edges have advantages when information is defined on graph edges.  Thus,  it is crucial to consider heterogeneous graphs where multiple types of features are defined on nodes,  edges,  and etc.  This is particularly suitable for brain functional data.

This study develops a novel heterogeneous graph convolutional neural network (HGCNN) simultaneously learning both nodes’ and edges’ functional features from fMRI data for predicting cognition or mental disorders. 
The HGCNN is designed to learn 1) nodes’ features from their neighborhood nodes’ features based on the topological connections of the nodes;  2) edges’ features from their neighborhood edges’ features based on the topological connections of the edges. To achieve these goals, the HGCNN considers a brain functional network as a simplex graph that allows characterizing node-node, node-edge, edge-edge, and higher-order topology. We develop a generic convolution framework by introducing the Hodge-Laplacian (HL) operator on the simplex graph and designing HL-spectral graph filters to aggregate features among nodes or edges based on their topological connections. In particular, this study takes advantage of spectral graph filters in \cite{defferrard2016convolutional,huang2021revisiting} and approximates HL-spectral graph filters using polynomials for 
spatial locations of these filters. We shall call our HGCNN as HL-HGCNN in the rest of the paper. Unlike the GNNs described above \cite{jiang2019censnet,zhao2022dynamic}, this study also introduces a simple graph pooling approach based on its topology such that the HL can be automatically updated for the convolution in successive layers, and the spatial dimension of the graph is reduced.  Hence,  the HL-HGCNN learns spectral filters along nodes, edges, or higher-dimensional simplex to extract brain functional features. 
 
We illustrate the use of the HL-HGCNN on fMRI time series and functional connectivity to predict general intelligence based on a large-scale adolescent cohort study (Adolescent Brain Cognitive Development (ABCD), n=7693). We also compare the HL-HGCNN with the state-of-art GNN techniques described above and demonstrate the outstanding performance of the HL-HGCNN.  Hence, this study proposes the following novel techniques:
\vspace{-0.2cm}
\begin{enumerate}
\item a generic graph convolution framework to smooth signals across nodes, edges, or higher-dimensional simplex;
\item spectral filters on nodes, edges, or higher-dimensional simplex via the HL operator;
\item HL-spectral filters with a spatial localization property via polynomial approximations;
\item a spatial pooling operator based on graph topology.
\end{enumerate}
\vspace{-0.6cm}

\section{Methods}
This study designs a heterogeneous graph convolutional neural network via the Hodge-Laplacian operator (HL-HGCNN) that can learn the representation of brain functional features at a node-level and an edge-level based on the graph topology.  In the following,  we will first introduce a generic graph convolution framework to design spectral filters on nodes and edges to learn node-level and edge-level brain functional representation based on its topology achieved via the HL operator.  We will introduce the polynomial approximation of the HL spectral filters to overcome challenges on 
spatial localization.  Finally,  we will define an efficient pooling operation based on the graph topology for the graph reduction and update of the HL operator.

\subsection{Learning Node-Level and Edge-Level Representation via the Hodge-Laplacian Operator}

In this study,  the brain functional network is characterized by a heterogeneous graph, $G=\{V, E\}$ with brain regions as nodes,  $V=\{v_i\}_{i=1}^n$,  and their connections as edges,  $E=\{e_{ij}\}_{i,j=1, 2, \cdots, n}$,  as well as functional time series defined on the nodes and functional connectivity defined on the edges.  This study aims to design convolutional operations for learning the representation of functional time series at nodes and the representation of functional connectivity at edges based on node-node and edge-edge connections (or the topology of graph $G$).

Mathematically,  nodes and edges are called $0-$ and $1-$dimensional simplex.  The topology of $G$ can be characterized by {\em boundary operator} $\boldsymbol{\partial}_k$.   $\boldsymbol{\partial}_1$ encodes how two 0-dimensional simplices,  or nodes,  are connecting to form a 1-dimensional simplex (an edge) \cite{edelsbrunner2000topological}. In the graph theory  \cite{lee2014hole},  
$\boldsymbol{\partial}_1$ can be represented as a traditional incidence matrix with size $n \times n(n-1)/2$,  where nodes are indexed over rows and edges are indexed over columns.  Similarly,  the second order boundary operator $\boldsymbol{\partial}_2$ encodes how 1-dimensional simplex,  or edges,  are connected to form the connections among 3 nodes (2-dimensional simplex or triangle).  

The goal of spectral filters is to learn the node-level representation of fMRI features from neighborhood nodes' fMRI features and the edge-level representation of fMRI features from neighborhood edges' fMRI features.  The neighborhood information of nodes and edges can be well characterized by the {\em boundary operators} $\boldsymbol{\partial}_k$ of graph $G$.  It is natural to incorporate the {\em boundary operators} of graph $G$ in the $k$-th Hodge-Laplacian (HL) operator defined as 
\begin{equation}
\label{eqn:hlopt}
\boldsymbol{\mathcal{L}}_k = \boldsymbol{\partial}_{k+1} \boldsymbol{\partial}_{k+1}^{\top} + \boldsymbol{\partial}_k^{\top} \boldsymbol{\partial}_k.
\end{equation}
When $k=0$,  the $0$-th  HL operator is 
\begin{equation}
\label{eqn:L0}
\boldsymbol{\mathcal{L}}_0= \boldsymbol{\partial}_1 \boldsymbol{\partial}_1 ^\top
\end{equation}
over nodes.  This special case is equivalent to the standard Graph Laplacian operator, $\boldsymbol{\mathcal{L}}_0 = \Delta$.
When $k=1$,  the $1$-st HL operator is defined over edges as
\begin{equation}
\label{eqn:L1}
\boldsymbol{\mathcal{L}}_1 = \boldsymbol{\partial}_{2} \boldsymbol{\partial}_{2}^{\top} + \boldsymbol{\partial}_1^{\top} \boldsymbol{\partial}_1  \ .
\end{equation} 

We can obtain orthonormal bases $\boldsymbol{\psi}_k^0, \boldsymbol{\psi}_k^1, \boldsymbol{\psi}_k^2, \cdots$ by solving eigensystem
$\boldsymbol{\mathcal{L}}_k\boldsymbol{ \psi}_k^j = \lambda_k^j  \boldsymbol{\psi}_k^j$.
We now consider an HL spectral filter $h$ with spectrum $h(\lambda_k)$ as
\begin{equation}\label{eq:h}
h(\cdot, \cdot)=\sum_{j=0}^{\infty}h(\lambda_k^j)\psi_k^j(\cdot) \psi_k^j(\cdot) .
\end{equation}
A generic form of spectral filtering of a signal $f$ on the heterogeneous graph $G$ can be defined as  
\begin{equation}\label{eq:hx}
 f'(\cdot) = h \ast f(\cdot)
=\sum_{j=0}^{\infty}h(\lambda_k^j) c_k^j\psi_k^j(\cdot) \ ,
\end{equation}
where $f(\cdot)= \sum_{j=0}^{\infty} c_k^j\psi_k^j(\cdot)$.   When $k=0$,  $f$ is defined on the nodes of graph $G$.  Eq.  (\ref{eq:hx}) indicates the convolution of a signal $f$ defined on $V$ with a filter $h$.  
 
Likewise,   when $k=1$,  $f$ is defined on the edges of graph $G$.   
Eq.  (\ref{eq:hx}) then indicates the convolution of a signal $f$ defined on $E$ with a filter $h$.  
Eq.  (\ref{eq:hx}) is generic that can be applied to smoothing signals defined on higher-dimensional simplices.  Nevertheless,  this study considers the heterogeneous graph only with signals defined on nodes and edges (0- and 1-dimensional simplices).   In the following,  we shall denote these two as "HL-node filtering" and "HL-edge filtering",  respectively.

\subsection{Laguerre Polynomial Approximation of the HL Spectral Filters}

The shape of spectral filters $h$ in Eq.  (\ref{eq:hx}) determines how many nodes or edges are aggregated in the filtering process.  Our goal of the HL-HGCNN is to design $h$ such as the representation at nodes and edges are learned through their neighborhood. This is challenging in the spectral domain since it requires $h(\lambda)$ with a broad spectrum.  
In this study,   we propose to approximate the filter spectrum $h(\lambda_k)$ in  Eq. (\ref{eq:hx}) as the expansion of Laguerre polynomials, $T_p$,  $p=0, 1, 2, \dots, P-1$,  such that
\begin{equation}
\label{eq:hlambda}
h(\lambda_k)=\sum_{p=0}^{P-1}\theta_p T_p(\lambda_k) \ ,
\end{equation}
where $\theta_p$ is the $p ^{th}$ expansion coefficient associated with the $p ^{th}$ Laguerre polynomial.  $T_p$ can be computed 
from the recurrence relation of $T_{p+1}(\lambda_k) = \frac{(2p+1-\lambda_k) T_{p}(\lambda_k)- pT_{p-1}(\lambda_k)}{p+1} $ with $T_0(\lambda_k) = 1$ and $T_1(\lambda_k) = 1 - \lambda_k$.

We can rewrite the convolution in Eq. (\ref{eq:hx}) as 
\begin{equation}
\label{eq:gx3}
f'(\cdot) = h \ast f(\cdot)=\sum_{p=0}^{P-1}\theta_p T_p(\boldsymbol{\mathcal{L}}_k) f(\cdot).
\end{equation}

Analog to the spatial localization property of the polynomial approximation of the graph Laplacian (the $0$-th HL) spectral filters \cite{huang2021revisiting,defferrard2016convolutional,wee2019cortical},  the Laguerre polynomial approximation of the $1$-st HL spectral filters can also achieve this localization property.  Assume two edges, $e_{ij}$ and $e_{mn}$, on graph $G$.  The shortest distance between $e_{ij}$ and $e_{mn}$ is denoted by $d_G(ij,mn)$ and computed as the minimum number of edges on the path connecting $e_{ij}$ and $e_{mn}$.  Hence,
  $  (\boldsymbol{\mathcal{L}}_1^P)_{e_{ij},e_{mn}} = 0 \quad if \quad  d_G(ij,mn) > P \ ,$
where $\boldsymbol{\mathcal{L}}_1^P$ denotes the $P$-th power of the $1$-st HL.  Hence,  the spectral filter represented by the $P$-th order Laguerre polynomials of the 1-st HL is localized within the $P$-hop edge neighborhood.   Therefore,  spectral filters in Eq.  (\ref{eq:hlambda}) have the property of spatial localization.  This proof can be extended to the $k$-th HL spectral filters.  In Section of Results \ref{sec:expts},  we will demonstrate this property using simulation data.

\vspace{-0.5cm}
\subsection{Topological Graph Pooling (TGPool)}
The pooling operation has demonstrated its effectiveness on grid-like image data \cite{yu2015multi}.  However,  spatial graph pooling is not straightforward,  especially for heterogeneous graphs.  This study introduces a generic topological graph pooling (TGPool) approach that  includes coarsening of the graph,  pooling of signals,  and an update of the Hodge-Laplacian operator.  For this,  we take an advantage of the one-to-one correspondence between the {\em boundary operators} and graph $G$ and define the three operations for pooling based on the {\em boundary operators}.  As the {\em boundary operators} encode the topology of the graph,  our graph pooling is topologically based.
\vspace{-0.5cm}
\begin{figure}[htbp!]
\begin{minipage}{0.6\textwidth}
\includegraphics[width=\textwidth]{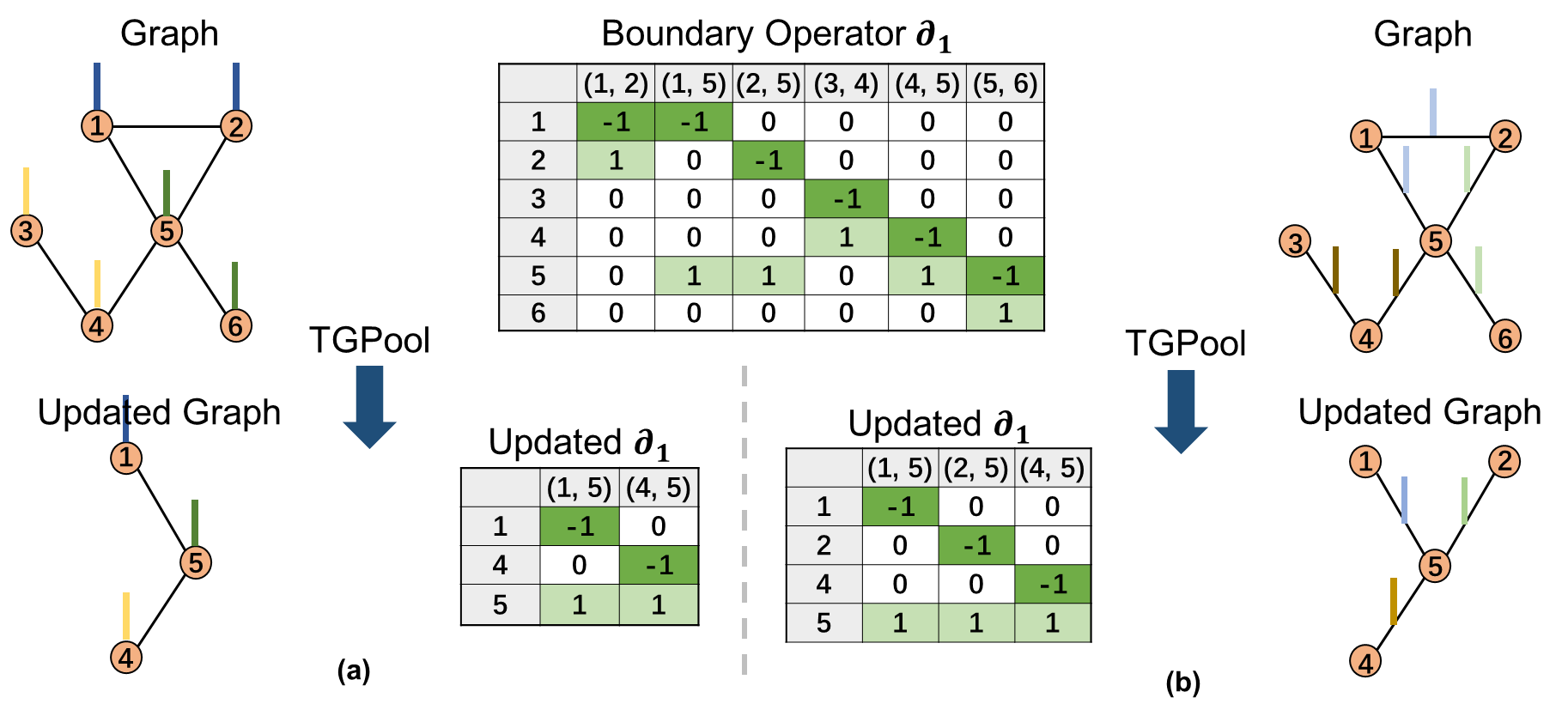}
\end{minipage}%
\begin{minipage}{0.4\textwidth}
\caption{Topological Graph Pooling (TGPool).  Panels (a) and (b) illustrate the topological graph pooling of (a) 0-dimensional (nodes) and (b) 1-dimensional (edges) simplices.  The color at each node or edge indicates features and their similarity across nodes or edges.} 
 \label{HL_pooling_illustration}
\end{minipage}
\end{figure}
\vspace{-0.8cm}

For graph coarsening,  we generalize the Graclus multilevel clustering algorithm \cite{dhillon2007weighted} to coarsen the $k-$dimensional simplices on graph $G$.  We first cluster similar $k-$dimensional simplices based on their associated features via local normalized cut.  At each coarsening level,  two neighboring $k-$dimensional simplices with maximum local normalized cut are matched until all $k-$dimensional simplices are explored \cite{shi2000normalized}.  A balanced binary tree is generated where each $k-$dimensional simplex has either one (i.e.,  singleton) or two child $k-$dimensional simplices.  Fake $k-$dimensional simplices are added to pair with those singletons.  The weights of $k+1-$dimensional simplices involving fake $k-$dimensional simplices are set as 0.  The pooling on this binary tree can be efficiently implemented as a simple 1-dimensional pooling of size 2.  Then,  two matched $k-$dimensional simplices are merged as a new $k-$dimensional simplex by removing the $k-$dimensional simplex with the lower degree and the $k+1-$dimensional simplices that are connected to this $k-$dimensional simplex.  To coarsen the graph,  we define a new {\em boundary operator} by deleting the corresponding rows and columns in the boundary operator and computing the HL operators via Eq.\ref{eqn:L0}.  Finally,  the signal of the new $k-$dimensional simplex  is defined as the average (or max) of the signals at the two $k-$dimensional simplices.  Fig.\ref{HL_pooling_illustration} illustrates the graph pooling of $0$-dimensional and $1$-dimensional simplices and the boundary operators of the updated graph after pooling.


\vspace{-0.3cm}
\subsection{Hodge-Laplacian Heterogeneous Graph Convolutional Neural Network (HL-HGCNN)}

\begin{figure}[htbp!]
\begin{center}
  \includegraphics[width=0.8\textwidth]{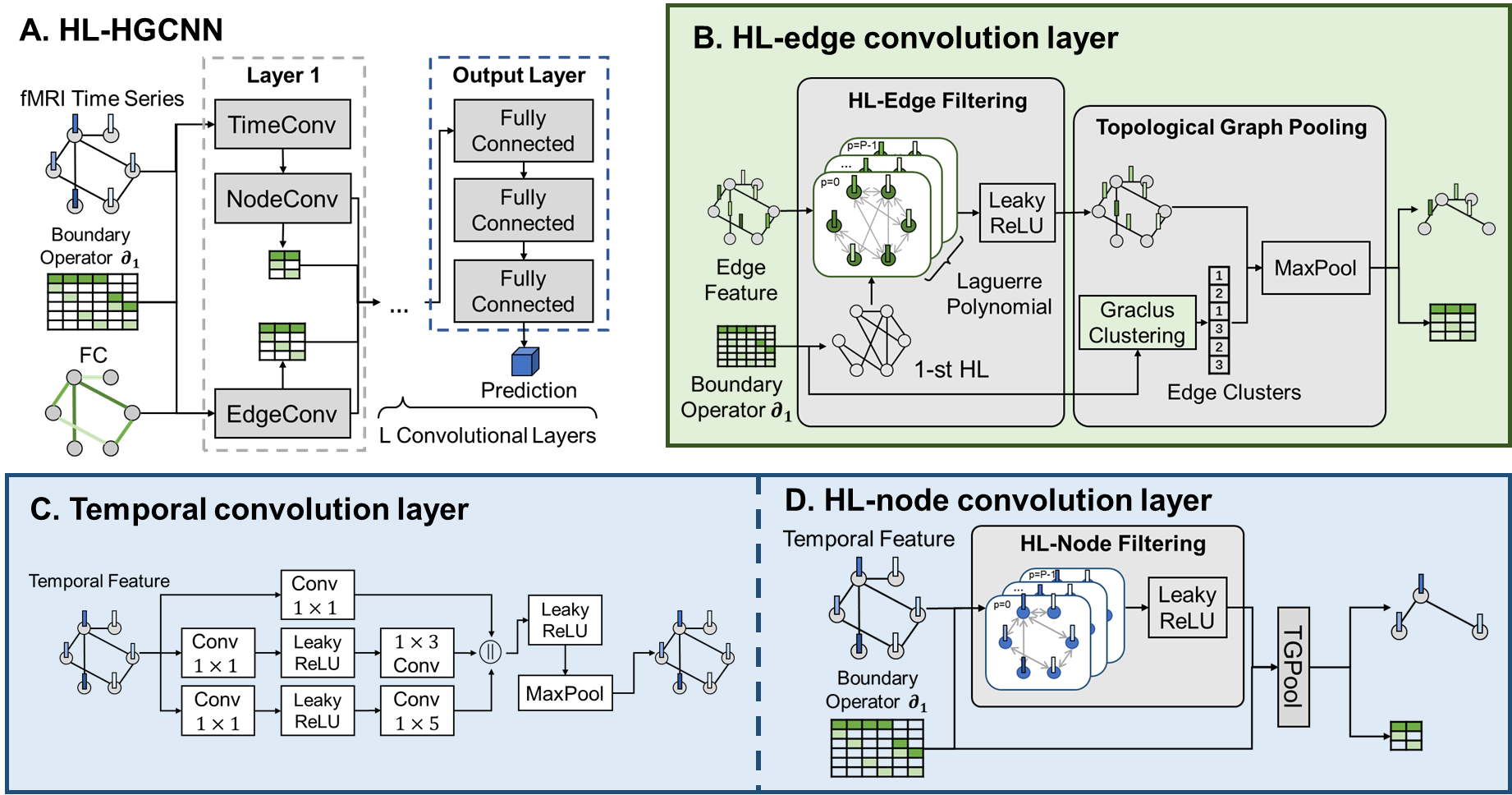}
  \caption{HL-HGCNN architecture.  Panel (A) illustrates the overall architecture of the HL-HGCNN model.  Panels (B-D) respectively show the architectures of the HL-edge, temporal, and HL-node convolutional layers.
}  
\label{network} 
\end{center}
\end{figure}

We design the HL-HGCNN with the temporal,  node,  and edge convolutional layers to learn temporal and spatial information of brain functional time series and functional connectivity.  Each layer includes the convolution, leaky rectified linear unit (leaky ReLU), and pooling operations.  Fig.\ref{network} illustrates the overall architecture of the HL-HGCNN model,  the temporal,  node,  and edge convolutional layers. 

\noindent {\bf Filters.} Denote $h_t$, $h_v$, $h_e$ to be temporal filters,  HL-node filters, HL-edge filters,  respectively.  $h_t$ is a simple 1-dimensional filter along the time domain with different kernel sizes to extract the information of brain functional time series at multiple temporal scales.  $h_v$ and $h_e$ are defined in Eq. (\ref{eq:hlambda}),  where $\theta_p$ are the parameters to be estimated in the HL-HGCNN.  As mentioned earlier, $P$ determines the kernel size of $h_v$ and $h_e$ and extracts the higher-order information of the brain functional time series and functional connectivity at multiple spatial scales. 

\noindent {\bf Leaky ReLU.} This study employs leaky rectified linear unit (ReLU) as an activation function, $\sigma$,  since negative functional time series and functional connectivity are considered biologically meaningful.  

\vspace{-0.5cm}
\vspace{-0.8cm}

\noindent {\bf Pooling.} In the temporal convolutional layer,  traditional 1-dimensional max pooling operations are applied in the temporal dimension of the functional time series.  In the edge and node convolutional layers,  TGPool is applied to reduce the dimension of the graph and the dimension of the node and edge signals.

\noindent {\bf Output Layer.} We use one more graph convolutional layer to translate the feature of each node or edge into a scalar.  Then, we concatenate the vectorized node and edge representations as the input of the output layer.  In this study,  the output layers contain fully-connected layers.

\vspace{-0.5cm}
\subsection{Implementation}
\label{sec:Implementation}
\noindent {\bf $\boldsymbol{\mathcal{L}}_0$ and $\boldsymbol{\mathcal{L}}_1$.}  Given a brain functional connectivity matrix,  we first build a binary matrix while the element in the connectivity matrix with its absolute value greater than a threshold is assigned as one,  otherwise zero.  We compute the boundary operator $\boldsymbol{\partial}_1$ with the size of the number of brain regions and the number of functional connectivities.   The $i$-th row of $\boldsymbol{\partial}_1$ encodes the functional connection of the $i$-th vertex and the $j$-th column of $\boldsymbol{\partial}_1$ encodes  how two vertices are connecting to form an edge \cite{edelsbrunner2000topological,edelsbrunner2002topological}.  Hence,  $\boldsymbol{\mathcal{L}}_0=\boldsymbol{\partial}_1 \boldsymbol{\partial}_1^{\top} $.  

According to Eq. (\ref{eqn:hlopt}),  the computation of $\boldsymbol{\mathcal{L}}_1$ involves the computation of $\boldsymbol{\partial}_2$ that characterizes the interaction of edges and triangles.  The brain functional connectivity matrix does not form a triangle simplex so the second order boundary operator $\boldsymbol{\partial}_2=0$.  Hence,  $\boldsymbol{\mathcal{L}}_1 =  \boldsymbol{\partial}_1^{\top} \boldsymbol{\partial}_1$.

\noindent {\bf Optimization. } We implement the framework in Python 3.9.13, Pytorch 1.12.1 and PyTorch Geometric 2.1.0 library. The HL-HGCNN is composed of two temporal,  node, and edge convolution layers with $\{8, 8\}$, $\{16, 1\}$, and $\{32, 32\}$ filters,  respectively.  The order of Laguerre polynomials for the 0-th and 1-st HL approximation is set to 3 and 4, respectively.  The output layer contains three fully connected layers with 256, 128 and 1 hidden nodes, respectively.  Dropout with 0.5 rate is applied to every layer and Leaky ReLU with a leak rate of 0.33 are used in all layers.  These model-relevant parameters are determined using greedy search. The HL-HGCNN model is trained using an NVIDIA Tesla V100SXM2 GPU with 32GB RAM by the ADAM optimizer with a mini-batch size of 32. The initial learning rate is set as 0.005 and decays by 0.95 after every epoch. The weight decay parameter was 0.005.

\vspace{-0.4cm}
\subsection{ABCD Dataset}
This study uses resting-state fMRI (rs-fMRI) images from the ABCD study that is an open-sourced and ongoing study on youth between 9–11 years old (\protect\url{https://abcdstudy.org/}). 
This study uses the same dataset of 7693 subjects and fMRI preprocessing pipeline stated in Huang et al. \cite{huang2022spatio}.  A node represents one of 268 brain regions of interest (ROIs) \cite{shen2017using} with its averaged time series as node features.  Each edge represents the functional connection between any two ROIs with the functional connectivity computed via Pearson’s correlation of their averaged time series as edge features.  General intelligence is defined as the average of 5 NIH Toolbox cognition scores, including Dimensional Change Card Sort, Flanker,  Picture Sequence Memory, List Sorting Working Memory, and Pattern Comparison Processing Speed \cite{akshoomoff2013viii}.  General intelligence ranges from  64 to 123 with mean and standard deviation of $95.3\pm 7.3$ among 7693 subjects.

\vspace{-0.4cm}
\section{Results}
\label{sec:expts}
\vspace{-0.2cm}
This section first demonstrates the spatial localization property of HL-edge filters in relation to the order of Laguerre polynomials via simulated data.  We then demonstrate the use of HL-edge filtering and its use in GNN for predicting fluid intelligence using the ABCD dataset.

\subsection{Spatial Localization of the HL-Edge Filtering via Laguerre Polynomial Approximations}
\begin{figure}[!htbp]
\begin{center}
\includegraphics[width=0.7\textwidth]{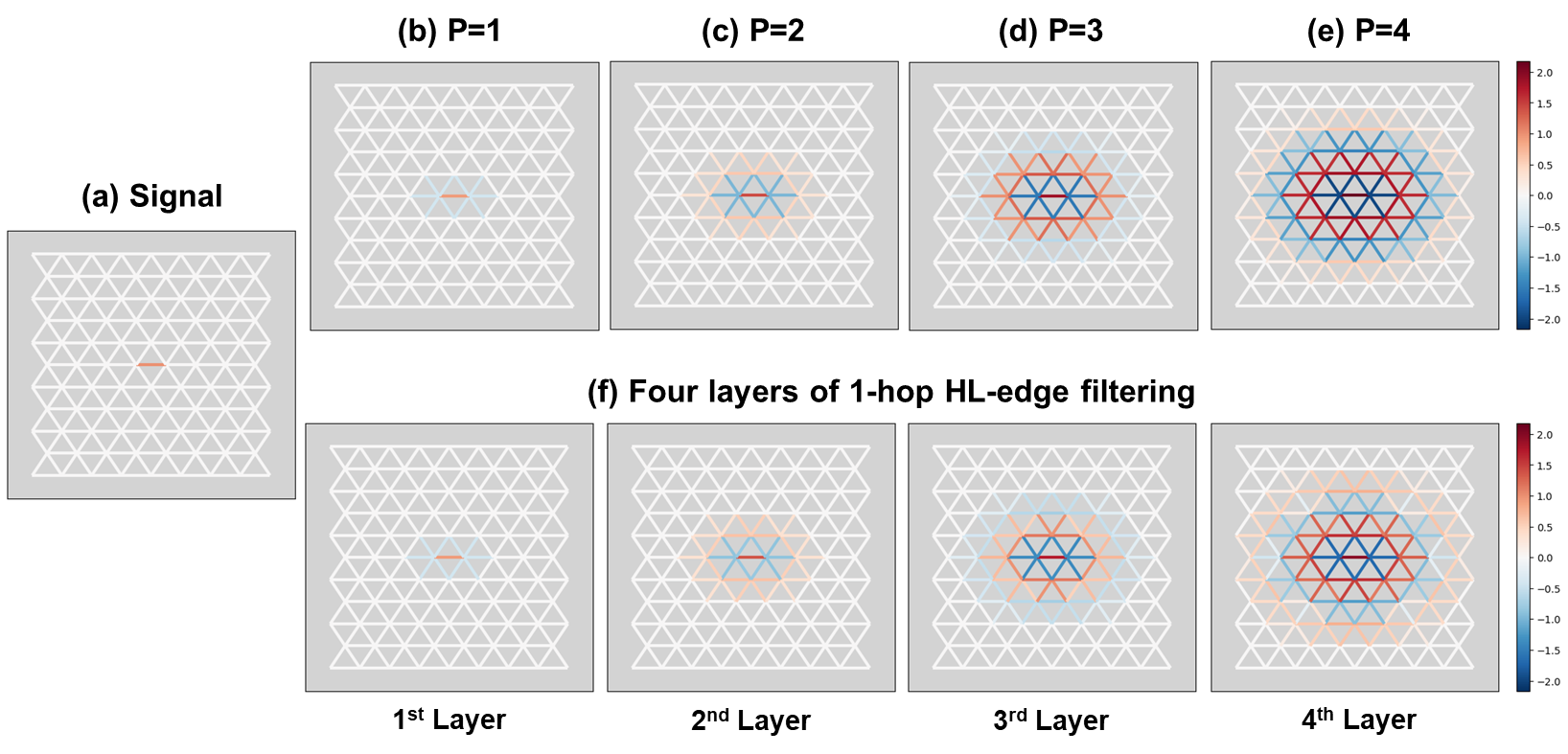}
\caption{Spatial localization of the HL-edge filtering.  Panel (a) shows the simulated signal only occurring at one edge.  Panels (b-e) show the signals filtered using the HL-edge filters with the $1^{st}$-,  $2^{nd}$-, $3^{rd}$-,  $4^{th}$-order Laguerre polynomial approximation,  respectively. Panel (f) illustrates the signals generated from the HL-edge convolution networks with 4 layers.  Each layer consists of the HL-filter approximated using the $1^{st}$-order Laguerre polynomial.}  
\label{HL_filter_size} 
\end{center}
\end{figure}

We illustrate the spatial location property of the HL-edge filtering by designing a pulse signal at one edge (Fig.  \ref{HL_filter_size} (a)) and smoothing it via the HL-edge filter.  When applying the HL-edge filter approximated via the $1^{st}$-,  $2^{nd}$-, $3^{rd}$-,  $4^{th}$-order Laguerre polynomials,  the filtered signals shown in Fig. \ref{HL_filter_size} (b-e) suggest that the spatial localization of the HL-edge filters is determined by the order of Laguerre polynomials.  This phenomenon can also be achieved using multi-layer HL-edge filters where each layer contains HL-edge filters approximated using the  $1^{st}$-order Laguerre polynomial (see Fig. \ref{HL_filter_size} (f)).
\vspace{-0.3cm}
\vspace{-1.5cm}
\subsection{HL-node vs.  HL-edge filters}
We aim to examine the advantage of the HL-edge filters over the HL-node filters when fMRI data by nature characterize edge information,  such as the functional connectivity.  When functional connectivities are defined at a node,  they form a vector of the functional connectivities related to this node.  In contrast,  by nature,  the functional connectivity represents the functional connection strength of two brain regions (i.e.,  edge).  Hence,  it is a scalar defined at an edge.  We design the HL-node network with the two HL-node convolutional layers (see in Fig. \ref{network}D) and the output layer with three fully connected layers.  Likewise,  the HL-edge network with the two HL-edge convolutional layers (see Fig. \ref{network}B) and the output layer with three fully connected layers.   We employ five-fold cross-validation six times to evaluate the prediction accuracy between predicted and actual general intelligence based on root mean square error (RMSE).  Table \ref{comparison_table} shows that the HL-edge network has smaller RMSE and performs better than the HL-node network ($p=1.51\times10^{-5}$ ).  This suggests the advantage of the HL-edge filters when features by nature characterize the weights of edges.

\subsection{Comparisons with existing GNN methods }
We now compare our models with the existing state-of-art methods stated above in terms of the prediction accuracy of general intelligence using the ABCD dataset.  The first experiment is designed to compare the performance of the HL-node network with that GAT \cite{hu2021gat}, BrainGNN \cite{li2020braingnn}, and dGCN \cite{zhao2022dynamic}.  We adopt the architecture of BrainGNN and dGCN from Li et al.  \cite{li2020braingnn} and \cite{zhao2022dynamic} as both methods were used for fMRI data.  The GAT is designed with two graph convolution layers,  each consisting of 32 filters and 2-head attention,  which is determined via greedy search as implemented in our model.   The functional connectivity vector of each region is used as input features.  Table  \ref{comparison_table} suggested that the HL-node network performs better than the GAT ($p=0.0468$) and BrainGNN (p=0.0195),  and performs equivalently with dGCN ($p=0.0618$). 

\vspace{-0.95cm}
\begin{table}[htbp!]
\centering
\caption{General intelligence prediction accuracy based on root mean square error (RMSE).  $p$-value is obtained from two-sample $t$-tests examining the performance of each method in reference to the proposed HL-HGCNN.
}
\scalebox{0.9}{
\begin{tabular}{c|c| p{2cm}<{\centering} p{2cm}<{\centering}}
\toprule[1pt]
& \textbf{GNN model} & \textbf{RMSE } & \textbf{$p$-value} \\
\hline
\multirow{4}{*}{GNN with node filtering} & \textbf{HL-Node network (ours)} & $7.134\pm 0.011$ &  $4.01\times10^{-6}$ \\ 
&\textbf{GAT} \cite{hu2021gat} & $7.165\pm 0.020$ & $1.91\times10^{-5}$ \\ 
&\textbf{BrainGNN}\cite{li2020braingnn} & $7.144\pm 0.013$ & $1.51\times10^{-6}$ \\ 
&\textbf{dGCN}\cite{li2020braingnn} & $7.151\pm 0.012$ & $9.83\times10^{-6}$ \\ 
\hline
\multirow{3}{*}{GNN with edge filtering} & \textbf{HL-Edge network (ours)} & $7.009\pm 0.012$ &  $ 2.48\times10^{-2}$ \\ 
& \textbf{BrainNetCNN}\cite{kawahara2017brainnetcnn} & $7.118\pm0.016$ &  $5.34\times10^{-6}$ \\
& \textbf{Hypergraph NN}\cite{jo2021edge} & $7.051\pm0.022$ & $3.74\times10^{-5}$ \\ 
\hline
GNN with node and edge filtering & \textbf{HL-HGCNN (ours)} & \textbf{6.972$\pm$0.015} &  - \\ 
\bottomrule[1.5pt]
\end{tabular}}
\label{comparison_table}
\end{table}

\vspace{-1.55cm}
\begin{figure}[htbp!]
\begin{minipage}{0.48\textwidth}
\includegraphics[width=\textwidth]{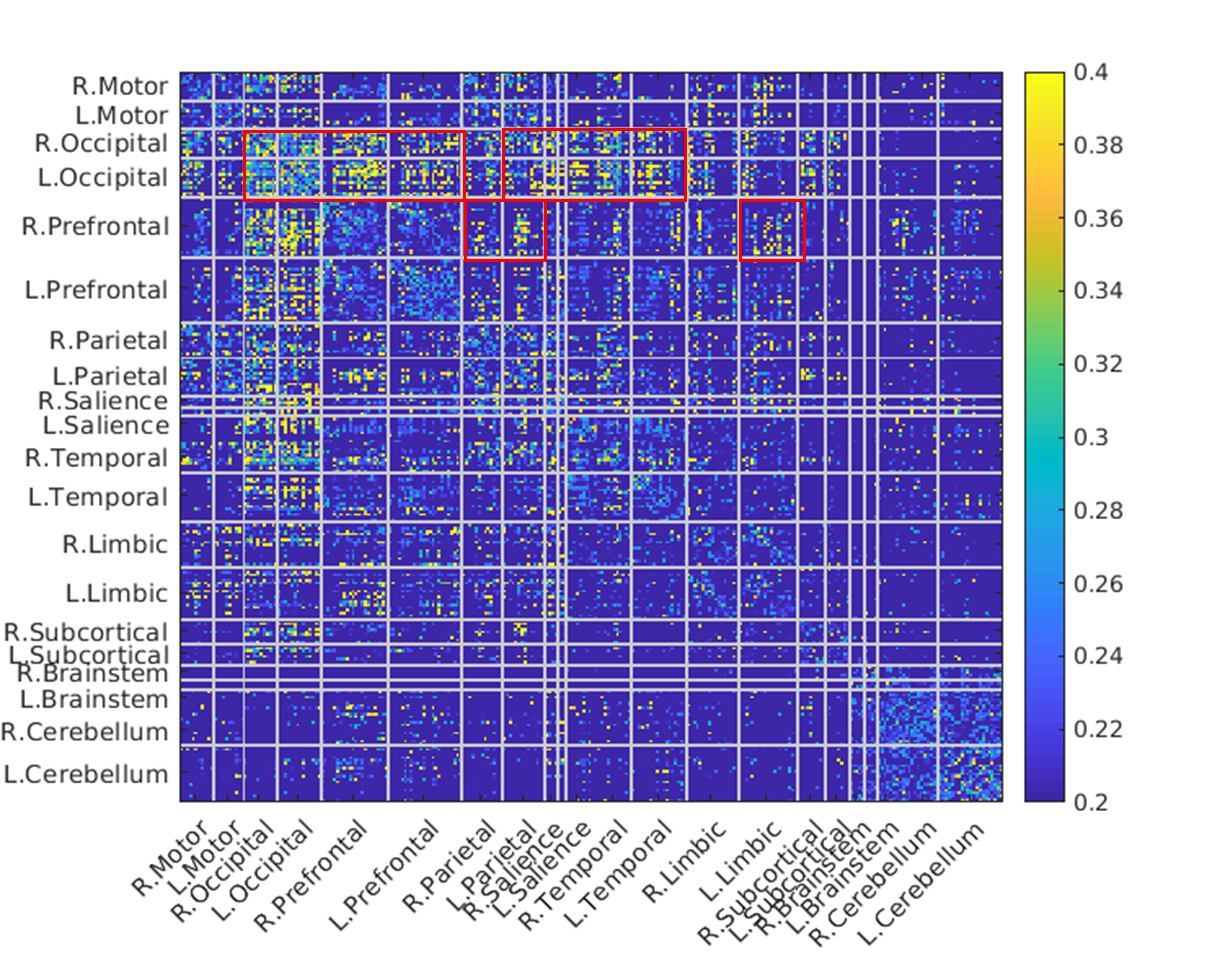}
\end{minipage}%
\begin{minipage}{0.4\textwidth}
\caption{The saliency map of the brain functional connectivity.  Red boxes highlight brain networks with higher weights, indicating greater contributions to the prediction of general intelligence.}
\label{cam}
\end{minipage}
\end{figure}
\vspace{-0.3cm}
The second experiment compares the HL-edge network with BrainNetCNN \cite{kawahara2017brainnetcnn} and Hypergraph NN \cite{jo2021edge}.  The Hypergraph NN comprises two graph convolution layers with 32 filters and one hypercluster layer after the first graph convolution layer.  The BrainNetCNN architecture follows the design in \cite{kawahara2017brainnetcnn}.  Table  \ref{comparison_table} shows that the HL-edge network has smaller RMSE and performs better than the BrainNetCNN ($p=4.49\times10^{-5}$) and Hypergraph NN ($p=0.0269$).   

Finally,  our HL-HGCNN integrates heterogeneous types of fMRI data at nodes and edges.  Table  \ref{comparison_table} shows that the HL-HGCNN performs the best compared to all the above methods (all $p<0.03$).

\vspace{-0.5cm}
\subsection{Interpretation}
We use the graph representation of the final edge convolution layer of the HL-HGCNN to compute the saliency map at the connectivity level. The group-level saliency map is computed by averaging the saliency maps across all the subjects in the dataset. The red boxes in Fig.\ref{cam} highlight the functional connectivities of the occipital regions with the prefrontal,  parietal,  salience, and temporal regions that most contribute to general intelligence.  Moreover,  our salience map also highlights the functional connectivities of the right prefrontal regions with bilateral parietal regions, which is largely consistent with existing findings on neural activities in the frontal and parietal regions \cite{jung_iq_2007,song_iq-2008}. 

\vspace{-0.4cm}
\section{Conclusion}
\vspace{-0.2cm}
This study proposes a novel HL-HGCNN on fMRI time series and functional connectivity for predicting cognitive ability.  Our experiments demonstrate the spatial localization property of HL spectral filters approximated via Laguerre polynomials.  Moreover,  our HL-node,  HL-edge,  and HL-HGCNN perform better than the existing state-of-art methods for predicting general intelligence,  indicating the potential of our method for future prediction and diagnosis based on fMRI.  Nevertheless,  more experiments on different datasets are needed to further validate the robustness of the proposed model. Our method provides a generic framework that allows learning heterogeneous graph representation on any dimensional simplices,  which can be extended to complex graph data.  The HL-HGCNN model offers an opportunity to build high-order functional interaction among multiple brain regions,  which is our future research direction.

\noindent {\bf Acknowledgements.}
This research/project is supported by the Singapore Ministry of Education (Academic research fund Tier 1) and A*STAR (H22P0M0007).  Additional funding is provided by the National Science Foundation MDS-2010778, National Institute of Health R01 EB022856, EB02875. 
This research was also supported by the A*STAR Computational Resource Centre through the use of its high-performance computing facilities.

\bibliographystyle{splncs03}
\bibliography{ref.gcnn,ref_fmri,deeplearning}

\begin{thebibliography}{10}
\providecommand{\url}[1]{\texttt{#1}}
\providecommand{\urlprefix}{URL }

\bibitem{akshoomoff2013viii}
Akshoomoff, N., Beaumont, J.L., Bauer, P.J., Dikmen, S.S., Gershon, R.C.,
  Mungas, D., Slotkin, J., Tulsky, D., Weintraub, S., Zelazo, P.D., et~al.:
  {VIII. NIH Toolbox Cognition Battery (CB}): composite scores of crystallized,
  fluid, and overall cognition. Monogr. Soc. Res. Child Dev.  78(4),  119--132
  (2013)

\bibitem{bruna2013spectral}
Bruna, J., Zaremba, W., Szlam, A., LeCun, Y.: Spectral networks and locally
  connected networks on graphs. arXiv preprint arXiv:1312.6203  (2013)

\bibitem{cui2022interpretable}
Cui, H., Dai, W., Zhu, Y., Li, X., He, L., Yang, C.: Interpretable graph neural
  networks for connectome-based brain disorder analysis. In: International
  Conference on Medical Image Computing and Computer-Assisted Intervention. pp.
  375--385. Springer (2022)

\bibitem{defferrard2016convolutional}
Defferrard, M., Bresson, X., Vandergheynst, P.: Convolutional neural networks
  on graphs with fast localized spectral filtering. In: Advances in Neural
  Information Processing Systems. pp. 3844--3852 (2016)

\bibitem{dhillon2007weighted}
Dhillon, I.S., Guan, Y., Kulis, B.: Weighted graph cuts without eigenvectors a
  multilevel approach. IEEE transactions on pattern analysis and machine
  intelligence  29(11),  1944--1957 (2007)

\bibitem{edelsbrunner2000topological}
Edelsbrunner, H., Letscher, D., Zomorodian, A.: Topological persistence and
  simplification. In: Proceedings 41st annual symposium on foundations of
  computer science. pp. 454--463. IEEE (2000)

\bibitem{edelsbrunner2002topological}
Edelsbrunner, H., Letscher, D., Zomorodian, A.: Topological persistence and
  simplification. Discrete Comput Geom  28,  511--533 (2002)

\bibitem{glover2011overview}
Glover, G.H.: Overview of functional magnetic resonance imaging. Neurosurgery
  Clinics  22(2),  133--139 (2011)

\bibitem{hu2021gat}
Hu, J., Cao, L., Li, T., Dong, S., Li, P.: Gat-li: a graph attention network
  based learning and interpreting method for functional brain network
  classification. BMC bioinformatics  22(1),  1--20 (2021)

\bibitem{huang2021revisiting}
Huang, S.G., Chung, M.K., Qiu, A.: Revisiting convolutional neural network on
  graphs with polynomial approximations of laplace--beltrami spectral
  filtering. Neural Computing and Applications  33(20),  13693--13704 (2021)

\bibitem{huang2022spatio}
Huang, S.G., Xia, J., Xu, L., Qiu, A.: Spatio-temporal directed acyclic graph
  learning with attention mechanisms on brain functional time series and
  connectivity. Medical Image Analysis  77,  102370 (2022)

\bibitem{jiang2019censnet}
Jiang, X., Ji, P., Li, S.: Censnet: Convolution with edge-node switching in
  graph neural networks. In: IJCAI. pp. 2656--2662 (2019)

\bibitem{jo2021edge}
Jo, J., Baek, J., Lee, S., Kim, D., Kang, M., Hwang, S.J.: Edge representation
  learning with hypergraphs. Advances in Neural Information Processing Systems
  34,  7534--7546 (2021)

\bibitem{jung_iq_2007}
Jung, R.E., Haier, R.J.: The parieto-frontal integration theory (p-fit) of
  intelligence: converging neuroimaging evidence. Behav. brain Sci.  30,
  135–154 (2007)

\bibitem{kawahara2017brainnetcnn}
Kawahara, J., Brown, C.J., Miller, S.P., Booth, B.G., Chau, V., Grunau, R.E.,
  Zwicker, J.G., Hamarneh, G.: {BrainNetCNN}: Convolutional neural networks for
  brain networks; towards predicting neurodevelopment. NeuroImage  146,
  1038--1049 (2017)

\bibitem{lee2014hole}
Lee, H., Chung, M.K., Kang, H., Lee, D.S.: Hole detection in metabolic
  connectivity of alzheimer's disease using k-laplacian. In: 17th International
  Conference on Medical Image Computing and Computer-Assisted Intervention,
  MICCAI 2014. pp. 297--304. Springer Verlag (2014)

\bibitem{li2020braingnn}
Li, X., Duncan, J.: {BrainGNN}: Interpretable brain graph neural network for
  {fMRI} analysis. bioRxiv  (2020)

\bibitem{shen2017using}
Shen, X., Finn, E.S., Scheinost, D., Rosenberg, M.D., Chun, M.M., Papademetris,
  X., Constable, R.T.: Using connectome-based predictive modeling to predict
  individual behavior from brain connectivity. Nature Protocols  12(3),
  506--518 (2017)

\bibitem{shi2000normalized}
Shi, J., Malik, J.: Normalized cuts and image segmentation. IEEE Transactions
  on pattern analysis and machine intelligence  22(8),  888--905 (2000)

\bibitem{song_iq-2008}
Song, M., Zhou, Y., Li, J., Liu, Y., Tian, L., Yu, C., Jiang, T.: Brain
  spontaneous functional connectivity and intelligence. Neuroimage  41,
  1168--1176 (2008)

\bibitem{wee2019cortical}
Wee, C.Y., Liu, C., Lee, A., Poh, J.S., Ji, H., Qiu, A., Initiative, A.D.N.:
  Cortical graph neural network for {AD} and {MCI} diagnosis and transfer
  learning across populations. NeuroImage: Clinical  23,  101929 (2019)

\bibitem{yu2015multi}
Yu, F., Koltun, V.: Multi-scale context aggregation by dilated convolutions.
  arXiv preprint arXiv:1511.07122  (2015)

\bibitem{zhao2022dynamic}
Zhao, K., Duka, B., Xie, H., Oathes, D.J., Calhoun, V., Zhang, Y.: A dynamic
  graph convolutional neural network framework reveals new insights into
  connectome dysfunctions in adhd. Neuroimage  246,  118774 (2022)

\end{thebibliography}

\end{document}